\documentclass[runningheads]{llncs}

\usepackage{eccv}

\usepackage{eccvabbrv}

\usepackage{epsfig}
\usepackage{geometry}
\usepackage{graphicx}
\usepackage{amsmath}
\usepackage{amssymb}
\usepackage{booktabs}
\usepackage{bm}
\usepackage{amsfonts}
\usepackage{mathrsfs}
\usepackage{pifont}
\usepackage{float}
\usepackage{multicol}
\usepackage{multirow}
\usepackage{adjustbox}
\usepackage{wrapfig}
\usepackage{array}
\usepackage[table]{xcolor}

\usepackage[accsupp]{axessibility}  

\usepackage[pagebackref,breaklinks,colorlinks,citecolor=eccvblue]{hyperref}
\usepackage[capitalize]{cleveref}
\crefname{section}{Sec.}{Secs.}
\Crefname{section}{Section}{Sections}
\Crefname{table}{Table}{Tables}
\crefname{table}{Tab.}{Tabs.}

\usepackage{orcidlink}
\newcommand{\modelname}{\textit{CompoSIA}\xspace}

\makeatletter
\newcommand{\printfnsymbol}[1]{%
        \textsuperscript{\@fnsymbol{#1}}%
}
\newcommand{\equalcontrib}{\textsuperscript{*}}
\newcommand{\corrauthor}{\textsuperscript{†}}
\newcommand{\projectlead}{\textsuperscript{‡}}
\makeatother

\begin{document}

\title{Composing Driving Worlds through Disentangled Control for Adversarial Scenario Generation}


\author{
\parbox{0.78\textwidth}{
\centering
Yifan Zhan\inst{1}\equalcontrib
\quad
Zhengqing Chen\inst{2}\equalcontrib\projectlead
\quad
Qingjie Wang\inst{2}\equalcontrib
\\
Zhuo He\inst{3}
\quad
Muyao Niu\inst{1}
\quad
Xiaoyang Guo\inst{2}
\quad
Wei Yin\inst{2}
\\
Weiqiang Ren\inst{2}
\quad
Qian Zhang\inst{2}
\quad
Yinqiang Zheng\inst{1}\corrauthor
}
}
\authorrunning{Zhan et al.}
\titlerunning{Composing Driving Worlds through Disentangled Control}

\institute{
\small
$^{1}$The University of Tokyo \quad
$^{2}$Horizon Robotics \quad
$^{3}$University of Glasgow
\\
\textsuperscript{*}Equal contribution \quad
\textsuperscript{‡}Project lead \quad
\textsuperscript{†}Corresponding author
}

\maketitle
\begin{center}
\vspace{-2ex}
\begin{minipage}{0.83\linewidth}
\includegraphics[width=1.0\linewidth]{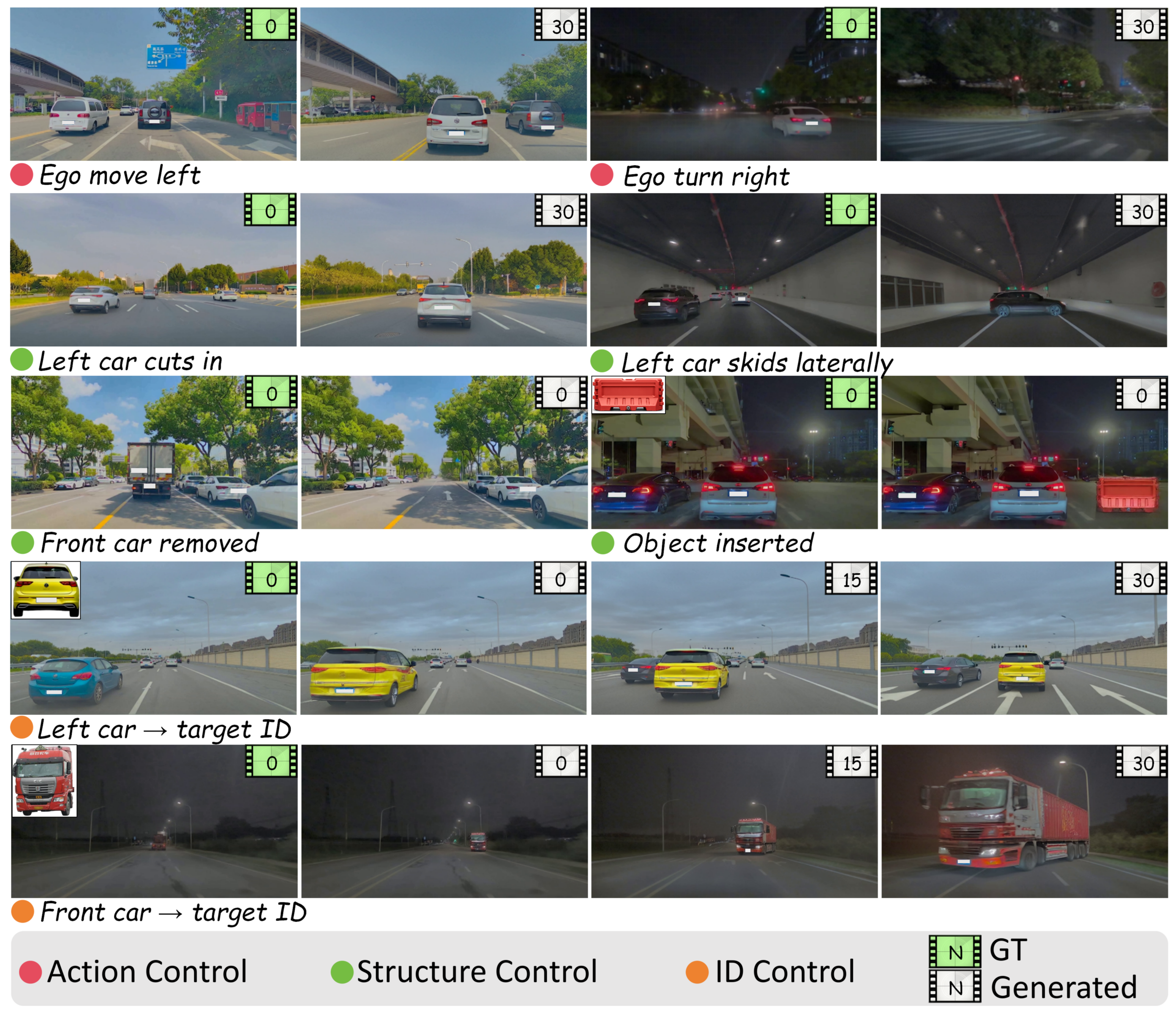}
\vspace{-4ex}
\captionof{figure}{\textbf{\modelname is a powerful simulator for synthesizing rare driving scenes.} While existing driving world models focus on generating, we enables explicit control over structure, identity, and ego action.}
\vspace{-2ex}
\label{fig: teaser}
\end{minipage}
\end{center}
\setlength{\parindent}{2em}

\begin{abstract}

A major challenge in autonomous driving is the ``long tail'' of safety-critical edge cases, which often emerge from unusual combinations of common traffic elements. 
Synthesizing these scenarios is crucial, yet current controllable generative models provide incomplete or entangled guidance, preventing the independent manipulation of scene structure, object identity, and ego actions.
We introduce \modelname, a compositional driving video simulator that disentangles these traffic factors, enabling fine-grained control over diverse adversarial driving scenarios.
To support controllable identity replacement of scene elements, we propose a noise-level identity injection, allowing pose-agnostic identity generation across diverse element poses, all from a single reference image.
Furthermore, a hierarchical dual-branch action control mechanism is introduced to improve action controllability.
Such disentangled control enables adversarial scenario synthesis—systematically combining safe elements into dangerous configurations that entangled generators cannot produce.
Extensive comparisons demonstrate superior controllable generation quality over state-of-the-art baselines, 
with a 17\% improvement in FVD for identity editing and reductions of 30\% and 47\% in rotation and translation errors for action control.
Furthermore, downstream stress-testing reveals substantial planner failures: across editing modalities, the average collision rate of $\mathtt{3s}$ increases by 173\%.
See project page at \href{https://github.com/Yifever20002/CompoSIA}{https://github.com/Yifever20002/CompoSIA}.
\keywords{Autonomous Driving \and Diffusion Models \and Controllable Generation}
\end{abstract}


\section{Introduction}
\label{sec:Introduction}

End-to-end autonomous driving systems rely on large-scale video data to learn unified representations for perception, prediction, and planning~\cite{ren2024diffusion, liao2025diffusiondrive, chen2024vadv2, li2024hydra}.
However, existing driving benchmarks such as nuScenes~\cite{caesar2020nuscenes} and Waymo~\cite{sun2020scalability} exhibit long-tail distributions, where rare scenarios often arise from unusual combinations of otherwise common traffic factors.
These adversarial samples are severely underrepresented, limiting scenario diversity and hindering the robustness of autonomous systems under complex driving situations.

To enable adversarial construction (\eg, a sudden truck cut-in forcing hard braking), a generative model~\cite{rombach2022high, peebles2023scalable} must support fine-grained control of scene structure, identity, and ego actions.
However, as summarized in \cref{tab:comparison}, existing generative approaches expose substantial limitations in 
controllability.
Video editing methods like DriveEditor~\cite{liang2025driveeditor} struggles to synthesize novel-view videos, while ego-action-based ReCamMaster~\cite{bai2025recammaster} fails with element-wise location and identity control.
More unified driving world models such as MagicDrive-V2~\cite{gao2025magicdrive} inject multiple conditions through shared pathways, leading to entangled generation and noticeable quality degradation with editing.
Overall, prior methods either control only a subset of scene factors, or fail to disentangle multiple signals, making it difficult to intentionally construct valuable adversarial scenarios---like ``\textit{a puppeteer pulling tangled strings}''.


To address these limitations, we argue that disentangled control requires factor-specific signals to be injected at different levels of the diffusion process.
Building on this insight, we propose \textbf{\modelname} (\textbf{Compo}site \textbf{S}tructure, \textbf{I}dentity, and \textbf{A}ction), a diffusion-based video generation framework with three explicit and complementary conditioning branches for structure, identity, and ego action control.
This design enables fine-grained independent and compositional control, allowing \modelname to precisely steer adversarial scenario generation.


For element-wise control, each scene element is manipulated through explicit structure and identity cues that play fundamentally different roles: structure governs geometric layout and motion trajectories, while identity specifies semantic appearance.
In our formulation, a scene element (\eg, a surrounding vehicle or a traffic cone) is defined by a 3D bounding box sequence and a single reference image.
This design is edit-friendly, yet raises a practical challenge: how to robustly inject semantic identity from a single image without over-constraining geometry.
To address this, we construct replaced-clean clip pairs and train an identity-guided denoiser with identity injection at multiple noise levels.
At sampling time, identity strength is tuned by the injection timestep, enabling controllable identity control across diverse element poses and categories.

\begin{table}[t]
\caption{\textbf{Comparison of control capabilities across general and driving-specific models.} 
Through disentangled conditioning, \modelname enables more comprehensive and fine-grained control and editing over adversarial scenario generation.}
\label{tab:comparison}
\vspace{-2mm}
\centering
\scalebox{0.7}{
\begin{tabular}{
>{\centering\arraybackslash}p{0.2\textwidth}
>{\arraybackslash}p{0.25\textwidth} 
>{\centering}p{0.15\textwidth}
>{\centering}p{0.15\textwidth}
>{\centering}p{0.15\textwidth}
>{\centering\arraybackslash}p{0.15\textwidth}
}
\toprule
\multicolumn{1}{c}{\multirow{2}{*}{\textbf{Type}}} & \multicolumn{1}{c}{\multirow{2}{*}{\textbf{Method}}} & 
\multicolumn{4}{c}{\textbf{Control Modes}} \\

 &  & Structure & Element ID & Scene ID & Ego-Action \\
\midrule

\multirow{4}{*}{\begin{tabular}[c]{@{}c@{}}General \\ Controllable \\ Generation\end{tabular}}
& ReCamMaster~\cite{bai2025recammaster}
& & & & \ding{51} \\
& Wan-Move~\cite{chu2025wan}
& \ding{51} & & \ding{51} & \ding{51} \\
& LoRA-Edit~\cite{gao2025lora}
& & \ding{51} & \ding{51} & \\
& TTM~\cite{singer2025time}
& \ding{51} & \ding{51} & \ding{51} & \\
\midrule

\multirow{6}{*}{\begin{tabular}[c]{@{}c@{}}Driving \\  Model\end{tabular}}
& Vista~\cite{gao2024vista}
& & & \ding{51} & \ding{51}\\
& Drive-WM~\cite{wang2024driving}
& & & \ding{51} & \ding{51}\\
& Panacea~\cite{wen2024panacea}
& \ding{51} & & \ding{51} & \ding{51}\\
& GAIA-2~\cite{russell2025gaia}
& \ding{51} & & \ding{51} & \ding{51}\\
& DriveEditor~\cite{liang2025driveeditor}
& \ding{51} & \ding{51} &  & \\
& MagicDrive-V2~\cite{gao2025magicdrive}
& \ding{51} & & & \ding{51}\\
\cmidrule{2-6}
&\modelname (Ours)
& \ding{51} & \ding{51} & \ding{51} & \ding{51} \\
\bottomrule
\end{tabular}
}

\centering
\includegraphics[width=0.9\linewidth]{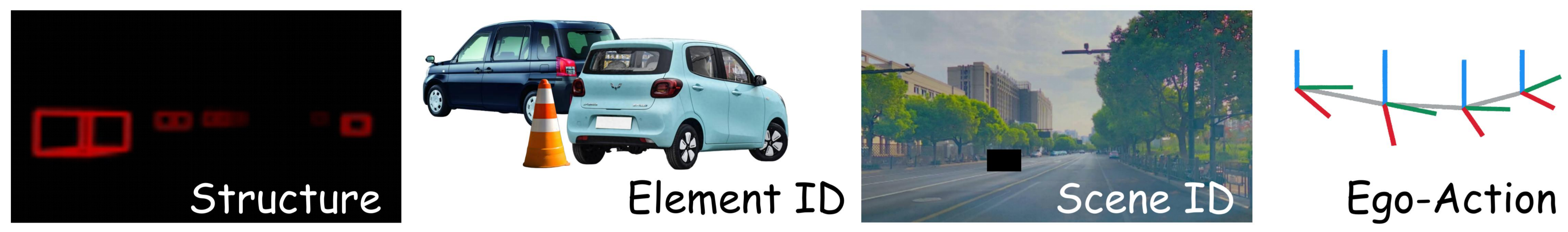}
\label{fig:intro_1}
\vspace{-3ex}
\end{table}


In contrast, ego action is naturally represented as a frame-wise continuous camera trajectory that governs global temporal dynamics.
To model such motion, we introduce a hierarchical dual-branch action conditioner that captures both local residual changes and global trajectory context.
The former introduces adaptive layer normalization (AdaLN)~\cite{peebles2023scalable} for fast early-stage learning, while the latter applies camera attention with PRoPE~\cite{li2025cameras} to provide accurate long-range guidance.
Together, the two branches accelerate convergence and improve the precision and stability of action-conditioned generation.


\modelname enables precise independent control and compositional editing of structure, identity, and ego action, turning driving world generation into a controllable simulator rather than a data synthesizer.
Our design allows deliberate construction of unusual factor combinations that are difficult to obtain from real-world logs and hard to produce with entangled synthesizer like DriveEditor~\cite{liang2025driveeditor} and MagicDrive-V2~\cite{gao2025magicdrive}.
Leveraging this capability, we design control policies to synthesize precise adversarial corner cases for planner stress-testing, exposing hidden failure modes beyond standard benchmarks.
Our contributions are three-fold:
\begin{itemize}

  \item [1)]
    a noise-level identity cue injection coupled with a tailored training strategy to recover consistent element identity from a single image, improving generalization across diverse element poses, categories, and in-the-wild scenes;
    
  \item [2)]
    a hierarchical dual-branch action conditioning scheme that models ego motion through complementary local residual modulation and global PRoPE attention, enabling faster convergence and more precise action control;

    \item[3)] 
    \modelname, a disentangled driving video simulator that enables independent and compositional control of structure, identity, and action, supporting systematic synthesis of adversarial scenarios for downstream stress-testing.

\end{itemize}


\section{Related Works}
\label{sec:Related Works}

\subsection{Video-based Driving World Models}

Recent progress in video generation has been largely driven by diffusion-based models~\cite{rombach2022high, peebles2023scalable}.
Foundational video generation models~\cite{brooks2024video, gao2025seedance, runwayml2024introducing, yang2024cogvideox, kong2024hunyuanvideo, ma2025step, wan2025wan, team2025longcat} have demonstrated strong capability in synthesizing high-quality, full-length videos conditioned on text prompts or reference images.
Autonomous driving scenarios require leveraging more complex sensor modalities to achieve higher-precision controllability.
A common line of work~\cite{yang2023bevcontrol, gao2023magicdrive, wen2024panacea, wang2024driving, wang2024drivedreamer, gao2024magicdrive3d, zhao2025drivedreamer, gao2025magicdrive} leverages geometric constraints, such as scene layouts or object-level spatial representations, to provide explicit structural guidance for generation.
Other approaches~\cite{hu2023gaia, lu2024wovogen, gao2024vista, guo2024infinitydrive, zhang2025epona, russell2025gaia} introduce ego-trajectory signals to better guide ego-vehicle behavior.
Semantic-guided driving scene generation remains underexplored, with existing approaches primarily relying on coarse-grained controls such as text prompts~\cite{hu2023gaia, russell2025gaia} or weather tags.

\subsection{Controllable Video Generation}

The injection of conditioning signals significantly improves the controllability of image and video generation models, including structure~\cite{ma2024follow, xing2024make, xing2024tooncrafter, zhou20243dis}, identity~\cite{hu2024animate, gong2023talecrafter, wang2025world, wang2024ms, chen2025xverse, shi2025consistcompose, xu2025contextgen}, scene flow~\cite{niu2024mofa, shi2024motion}, camera pose~\cite{bai2025recammaster, wang2025cinemaster, yu2024viewcrafter}, 3D point clouds~\cite{ren2025gen3c, yu2024wonderjourney, ma2025you, li2025vmem}, and audio signals~\cite{haji2025av, lee2022sound}.
Early UNet-based diffusion models rely on designs such as ControlNet~\cite{zhang2023adding} or IP-Adapter~\cite{ye2023ip} for conditioning injection.
The rise of diffusion transformers~\cite{peebles2023scalable} has introduced greater flexibility for conditioning injection.
Latent concatenation injects conditions into the data stream at fixed positions.
Cross-attention layers treat conditions as external context to provide global control over the generation process.
More recently, in-context conditioning~\cite{ju2025fulldit} has been explored, which places both the input and all conditioning signals into a single token sequence, allowing the model to reason over them jointly within a unified  framework.

\subsection{Image-based Identity Control and Editing}

Image-based generation has been extensively studied.
ID-Animator~\cite{he2024id} employs a ViT-based face adapter with cross-attention for identity conditioning.
SkyReels-A2~\cite{fei2025skyreels} further integrates multi-view identity information and prompt concepts into a temporally aware DiT model.
ConsisID~\cite{yuan2025identity} and MagicMe~\cite{ma2024magic} mitigate identity drift by introducing frequency-aware and dynamic attention, respectively.
However, for attention mechanisms, enforcing stronger identity constraints can hinder motion expressiveness, while dominant motion cues may distort identity.
TTM~\cite{singer2025time} performs identity-guided editing at the noise level using reference images, while training-free formulation makes it difficult to achieve identity control under accurate motion.
To preserve motion correctness and background consistency, identity guidance in autonomous driving~\cite{zeng2025rethinking, wang2025mirage, zeng2025rethinking, ljungbergh2025r3d2, li2025realistic, liang2025driveeditor, he2025langdrivectrl} is applied conservatively, operating on pose-aligned single image or image sequences for mild harmonization.
Such designs typically require time-consuming pose alignment and sacrifice the ability to generate novel scenes.


\section{Methods}
\label{sec:Methods}

\begin{figure}[t]
\centering
\includegraphics[width=1\linewidth]{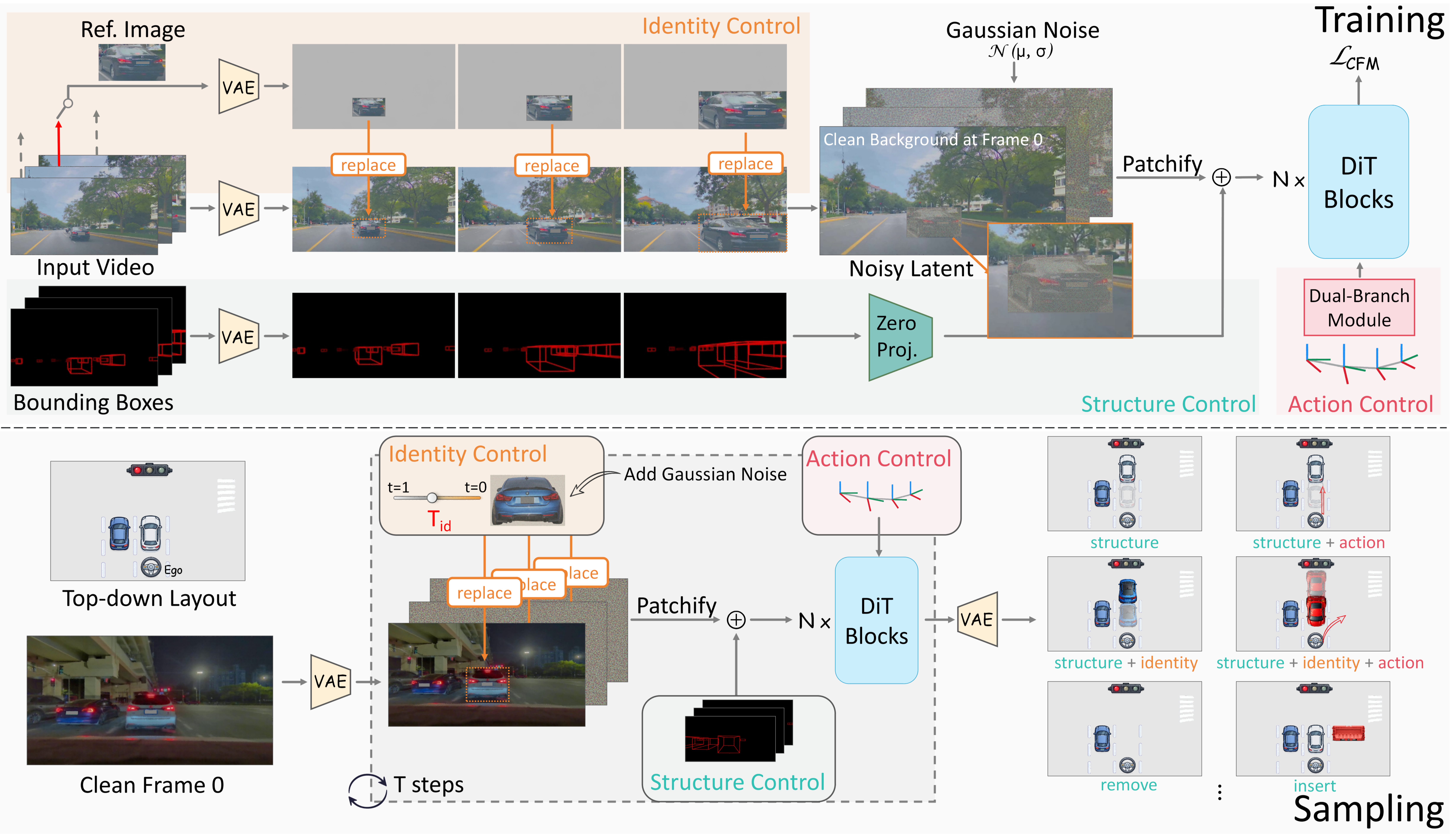}
\caption{\textbf{Overview of the \modelname framework.} During training (top), structure, identity, and ego-action signals are explicitly decomposed and injected into a Flow Matching–based DiT backbone in a disentangled yet compositional manner. During sampling (bottom), different combinations of conditions enable controllable driving video generation, supporting multiple editing applications.}
\label{fig:pipeline}
\end{figure}

This section presents the formulation and architectural design of our controllable driving world simulator, \modelname (see~\cref{fig:pipeline}). 
We first introduce the Flow Matching~\cite{albergo2022building, lipman2022flow, liu2022flow}-based DiT formulation that serves as the backbone of our framework.
Built upon this foundation, we describe how three categories of control signals—3D structure, identity, and ego action—are explicitly decomposed and injected during training under a disentangled yet compositional design principle. 
Finally, we detail the sampling procedure and show how different combinations of conditions enable structured and controllable video generation.

\subsection{Preliminary}

For efficient video generation, diffusion models are typically trained and sampled in latent spaces~\cite{rombach2022high, esser2024scaling, blattmann2023stable, brooks2024video, zheng2024open}, enabled by well-designed compression VAEs.
Specifically, a diffusion process that transforms samples $\boldsymbol{\epsilon}$ drawn from a prior distribution $p_1 = \mathcal{N}(\mathbf{0}, \mathbf{I})$ into latent data samples $z_{(0)}$ is formulated through a differentiable dynamical equation $\Theta$ as
\begin{equation}
\label{equ:flow_matching}
    \mathrm{d}z_{(t)} = v_\Theta(z_{(t)}, t)\,\mathrm{d}t, \quad t \in [0, 1].
\end{equation}

\modelname also adopts a VAE-diffusion framework, incorporating recent flow matching formulations~\cite{albergo2022building,
lipman2022flow, liu2022flow} and supervised with a $v$-prediction loss.
Intermediate states are constructed as
\begin{equation}
\label{equ:states}
    z_{(t)} = \sigma_t\,z_{(0)} + (1 - \sigma_t)\,\boldsymbol{\epsilon},
\end{equation}
and the objective is
\begin{equation}
    \mathcal{L}_{CFM} =
    \mathbb{E}_{\boldsymbol{\epsilon} \sim \mathcal{N}(\mathbf{0}, \mathbf{I})}
    \left\|
        v_\Theta(z_{(t)}, t) - (z_{(0)} - \boldsymbol{\epsilon})
    \right\|_2^2.
\end{equation}
where $\sigma$ denotes a predefined noise schedule.

\subsection{Structure Conditioning via Spatiotemporal Layouts}
\label{subsec:Structure Conditioning via Spatiotemporal Layouts}

To faithfully represent driving scenes, element-wise annotations of both dynamic and static elements (\eg, vehicles and traffic cones) are essential.
Incorporating such structured annotations into the generative model substantially enhances element controllability.
Given a video clip $x_{(0)} \in \mathbb{R}^{F \times H \times W \times 3}$, 
we obtain element-wise dynamic 3D bounding box annotations from video frames or LiDAR sensors.
Each element is represented by a bounding box sequence $\bm{b} \in \mathbb{R}^{F \times 7}$, encoding its 3D position, size, and orientation.
However, these spatial annotations lie in a 3D world coordinate system, which is mismatched with the 2D image domain of latent video generation.
To bridge this gap, rather than directly embedding $\bm{b}$, we first convert each box into a 3D polyline $\bm{p}$ and project it onto the image plane via camera intrinsics $K$ and extrinsics $E$, following
\begin{equation}
\bm{b}_{f} = \mathrm{Pinhole} (\bm{p}, K, E),
\end{equation}
where $\bm{b}_{f} \in \mathbb{R}^{F \times H \times W \times 3}$ is spatiotemporally aligned with $x_{(0)}$.

The diffusion process is performed in the latent space.
Given a VAE encoder $\mathcal{E}$, we obtain $z_{(0)} = \mathcal{E}(x_{(0)}) \in \mathbb{R}^{c \times f \times h \times w}$.
This latent is perturbed with noise according to the schedule $\sigma_t$ (~\cref{equ:states}).
It is then partitioned into patches of size $P$, which are flattened into a sequence of latent tokens $\bm{h}_{(t)} \in \mathbb{R}^{f \times s \times d}$ ($s = \frac{h}{P} \times \frac{w}{P}$, and $d$ is the channel dimension of latent tokens).

To incorporate the structure cue $\bm{b}_{f}$, we first encode it with $\mathcal{E}$ to obtain $z_{\bm{b}_{f}} \in \mathbb{R}^{c \times f \times h \times w}$.
The latent is further processed by a lightweight
convolutional adapter and reshaped into
layout tokens $\bm{h}_{\bm{b}_{f}} \in \mathbb{R}^{f \times s \times d}$. 
Structure condition is achieved through
a zero-initialized projection module
\begin{equation}
\bm{h}_{(t)} \leftarrow \bm{h}_{(t)} + f_{\text{zero}}\!\left(\bm{h}_{\bm{b}_{f}}\right).
\end{equation}

\subsection{Identity Conditioning via Noise Editing}

We aim to control the identity of a specific scene element in a video clip using only a single reference image.
This remains challenging for existing methods, as prior video editing works~\cite{liang2025driveeditor, ljungbergh2025r3d2, wang2025mirage} lack precise control over element pose and position, while attention-based conditioning alone provides insufficient identity supervision.
To this end, we formulate identity control as a restoration problem in the diffusion process: identity cue is injected at noisy stages, and the model is trained to restore a clean, geometrically consistent sequence.

To achieve this, we first construct training pairs using a \textit{select-and-repaint} strategy.
For each element in a video clip $x_{(0)}$, we randomly select one frame in which it appears and crop a reference image.
Next, for each element, we spatially align its reference image to the sequence of 2D bounding boxes (\ie, enclosing rectangles of projected 3D boxes $\bm{b}_{f}$) across the clip and paste it back frame by frame according to the projected locations, forming the identity cue condition $\bm{r}_f \in \mathbb{R}^{F \times H \times W \times 3}$.
This step also produces a reference mask $\bm{M}_{\bm{r}_f} \in \mathbb{R}^{F \times H \times W}$, indicating regions within the 2D bounding box sequences.
Now for each element, we construct a training pair consisting of the original natural video and a hard-anchored reference image sequence.

During training, we employ a hard-binding strategy at high-noise stages, replacing the corresponding regions of $x_{(0)}$ with $\bm{r}_f$ to enforce identity supervision.
Let ${z_{\bm{r}_f}}_{(0)} = \mathcal{E}(\bm{r}_f) \in \mathbb{R}^{c \times f \times h \times w}$ and $\bm{m}_{\bm{r}_f} \in \mathbb{R}^{f \times h \times w}$ denote the encoded identity cue and its downsampled mask. 
This injection can be formulated as
\begin{equation}
    {z_{\bm{r}_f}}_{(t)} = \sigma_t\,{z_{\bm{r}_f}}_{(0)} + (1 - \sigma_t)\,\boldsymbol{\epsilon},
\end{equation}
\begin{equation}
z_{(t)} \leftarrow \bm{m} \odot {z_{\bm{r}_f}}_{(t)} + (1 - \bm{m}) \odot z_{(t)},
\end{equation}
where
\begin{equation}
\bm{m} = \bm{m}_{\bm{r}_f} \cdot \mathbb{I}\,(t > T_{\text{id}}).
\end{equation}
Here, $\mathbb{I}\,(t > T_{\text{id}})$ indicates that the identity injection is active only at at high-noise stages.
We avoid hard replacement at low-noise stages, as it would disrupt the model’s generation pathway established during late-stage denoising.
This restoration-based formulation enables efficient and stable identity learning during training.
As we will detail later, when combined with our sampling strategy, it further achieves dual fidelity in both element identity and pose.


\subsection{Hierarchical Dual-Branch Action Conditioning}

\begin{wrapfigure}{r}{0.4\textwidth}
\centering
\vspace{-8ex}
\includegraphics[width=\linewidth]{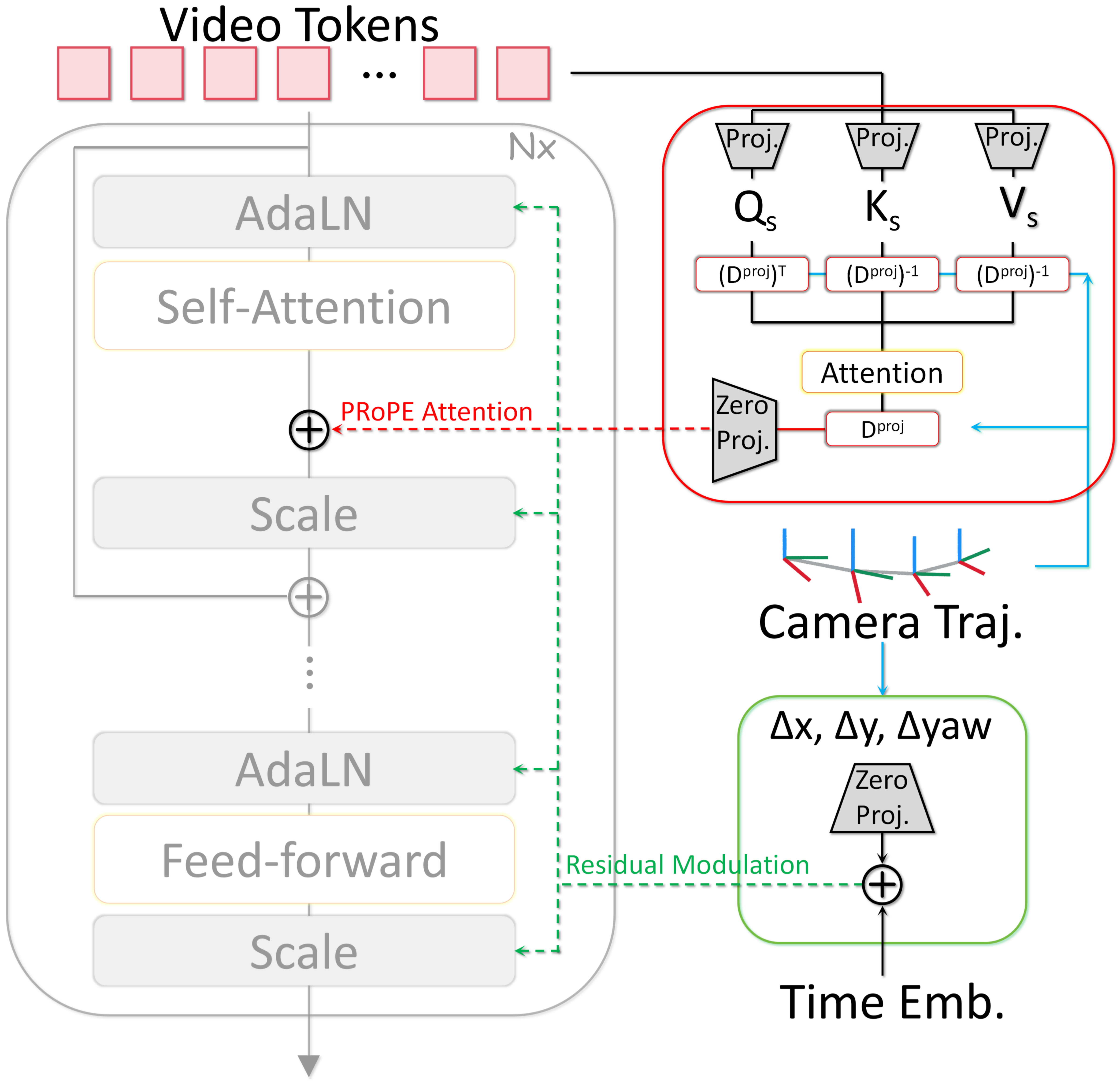}
\caption{\textbf{Design of hierarchical dual-branch action Conditioning.} Local residual modulation accelerates early convergence, while global PRoPE embedding improves overall accuracy.}
\label{fig:action}
\vspace{-6ex}
\end{wrapfigure}

Ego action is modeled as frame-wise and continuous camera trajectory.
In~\cref{fig:action}, our action conditioning adopts a hierarchical design with both local and global injection pathways.
We observe that local injection facilitates rapid convergence during the early stages of training, while global injection improves the overall accuracy and stability.

\noindent \textbf{Local Residual Modulation.}
Given a continuous trajectory of shape $F \times 4 \times 4$, we compute the relative transformation $\Delta \mathbf{T}_i = \mathbf{T}_{i}^{-1}\mathbf{T}_{i+1}$ ($\Delta \mathbf{T}_F=\Delta \mathbf{T}_{F-1}$), from which we extract $\bm{a}=(\Delta x, \Delta y, \Delta \text{yaw}) \in \mathbb{R}^{F \times 3}$ to serve as the residual signal.
This is injected through a gating mechanism implemented via adaptive layer normalization~\cite{peebles2023scalable} following $\!f_{\text{zero}}\!\left(\phi(\bm{a})\right) \in \mathbb{R}^{f \times 6 \times d}$, where $\phi(\cdot)$ denotes the sinusoidal frequency embedding and $f_{\text{zero}}(\cdot)$ is a zero-initialized projector.
The six channels are split into two groups of three, representing the pre-normalization shift, scale, and post-layer residual gate for the self-attention and feed-forward sublayers, respectively.
This local branch enables the model to quickly learn action control during early training.

\noindent \textbf{Global PRoPE Embedding.}
The local branch inherently loses precise motion information and camera intrinsic cues.
To improve the overall accuracy of action control, we embed the global trajectory with Projective Positional Encoding (PRoPE)~\cite{li2025cameras}.
Recall the original self attention computation as 
\begin{equation}
Attn_{v} = Attn(R^\top \odot Q, R^{-1} \odot K, V) ,
\end{equation}
where $R$ is the 3D RoPE~\cite{su2024roformer} and $\odot$ denotes the matrix-vector product.
We further compute
\begin{equation}
Attn_{c} = D^{proj} \odot Attn\!\left((D^{proj})^\top \odot Q_s,\; (D^{proj})^{-1} \odot K_s,\; (D^{proj})^{-1} \odot V_s\right).
\end{equation}
Here, $D^{proj}$ is derived from camera intrinsic and extrinsic parameters.

However, the 3D RoPE design in pre-trained video models and the camera projection design utilize different channel-wise structures.
This architectural discrepancy makes it difficult to directly merge relative camera pose embeddings (PRoPE) into the base model without disrupting its generative priors.
To address this, we design a lightweight PRoPE attention control branch initialized with zero-convolutions.
Furthermore, we observe that camera trajectory signals are inherently low-dimensional, for which a full-dimensional attention space is unnecessary.
To minimize computational overhead, we first project the video tokens into a lower-dimensional subspace (\eg, $1/8\times$) to compute reduced $(Q_s, K_s, V_s)$. 
The camera-conditioned attention is efficiently computed within this compact space and subsequently projected back to the original feature dimension, before being safely injected into the main attention branch via the zero-convolution layers.
The final attention output is computed as $Attn_v + f_{\text{zero}}(Attn_c),$
where $f_{\text{zero}}$ is a zero-initialized projector.
This global mechanism enables the model to achieve more precise action control.


\subsection{Training–Sampling Design for Condition Decoupling}
For the training modalities, we observe that structure cue in~\cref{subsec:Structure Conditioning via Spatiotemporal Layouts} can leak ego-action information during training (\eg, when other vehicles gradually move backward, it often indicates that the ego vehicle is moving forward rather than remaining static).
To prevent this implicit supervision from biasing the learning process, structure cue is always paired with action conditioning.  
Further experiments show that structure and identity cues do not conflict with each other and can be jointly trained.
In practice, we adopt a training modality ratio of 0.6:0.3:0.1 for [action], [structure, identity, action], and unconditional inputs, respectively.
The background region of the first frame is replaced with a clean latent~\cite{liu2025pusa}, and its denoising timestep is reset to zero to explicitly inject scene identity information.
Moreover, we additionally apply localized noise perturbation to the first frame during training to enhance the model's inpainting capability.

\begin{figure}[t]
\centering
\includegraphics[width=0.9\linewidth]{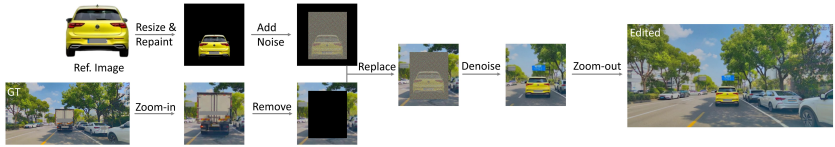}
\caption{\textbf{Identity editing for the first frame.} We preserve the background to anchor scene identity while editing the foreground. If the reference area is smaller than the original area, the intermediate region is treated as an inpainting area during denoising.}
\label{fig:edit}
\vspace{-3ex}
\end{figure}

During sampling, independent control of each condition becomes possible.
The structure condition can be edited separately (\eg, modifying the motion or presence of other elements).
When editing action alone, the structure condition must be explicitly re-projected to ensure that the motion of other elements remains unchanged.
The reference image for identity editing can be drawn from the dataset or from in-the-wild images, and may correspond to vehicles or other traffic elements.
We now provide a detailed description of the first-frame processing (\cref{fig:edit}) in our identity editing pipeline.
The background region is explicitly preserved as a clean latent to anchor scene identity.
For element identity, when the reference image is smaller than the original edited region, the area between the foreground and background is treated as an inpainting region during denoising.
Such inpainting configurations are explicitly exposed during training, enabling robust identity reconstruction under partial spatial replacement.
See~\cref{sec:Experiments} for more visualizations.


\section{Experiments}
\label{sec:Experiments}

\begin{table}[t]
\caption{\textbf{Quantitative comparison of different methods.} We evaluate multi-view scene following, identity control, and action control tasks.}
\label{tab:FVD}
\vspace{-2mm}
\centering
\scalebox{0.7}{
\begin{tabular}{
>{\centering\arraybackslash}p{0.15\textwidth}
>{\arraybackslash}p{0.25\textwidth} 
>{\centering}p{0.1\textwidth}
>{\centering\arraybackslash}p{0.25\textwidth}
}
\toprule
Type & Method & FVD$\downarrow$ & VBench Score (\%)$\uparrow$ 
\\
\midrule
\multirow{2}{*}{\begin{tabular}[c]{@{}c@{}}6v-Scene \\ Following\end{tabular}}
& MagicDrive-V2~\cite{gao2025magicdrive}
& 152.80 & 77.23 \\
&\cellcolor{gray!15}\modelname (Ours)
&\cellcolor{gray!15} \textbf{133.66} &\cellcolor{gray!15} \textbf{81.05}     \\
\midrule
\multirow{4}{*}{\begin{tabular}[c]{@{}c@{}}Identity \\ Control\end{tabular}}
& TTM~\cite{singer2025time}
& 231.17 & 75.16  \\
& LoRA-Edit~\cite{gao2025lora}
& 161.32 & 79.83   \\
& DriveEditor~\cite{liang2025driveeditor}
& 179.57 & 79.13  \\
&\cellcolor{gray!15}\modelname (Ours)
&\cellcolor{gray!15} \textbf{149.15} &\cellcolor{gray!15}  \textbf{80.30}     \\
\midrule
\multirow{4}{*}{\begin{tabular}[c]{@{}c@{}}Action \\ Control\end{tabular}}
& ReCamMaster~\cite{bai2025recammaster}
& 190.52 & 74.29   \\
& Vista~\cite{gao2024vista}
& 171.49 & 75.35   \\
& MagicDrive-V2~\cite{gao2025magicdrive}
& 279.61 & 73.44   \\
&\cellcolor{gray!15}\modelname (Ours)
&\cellcolor{gray!15} \textbf{137.21} &\cellcolor{gray!15} \textbf{80.79}    \\
\bottomrule
\end{tabular}
}
\vspace{1ex}

\centering
\includegraphics[width=0.9\linewidth]{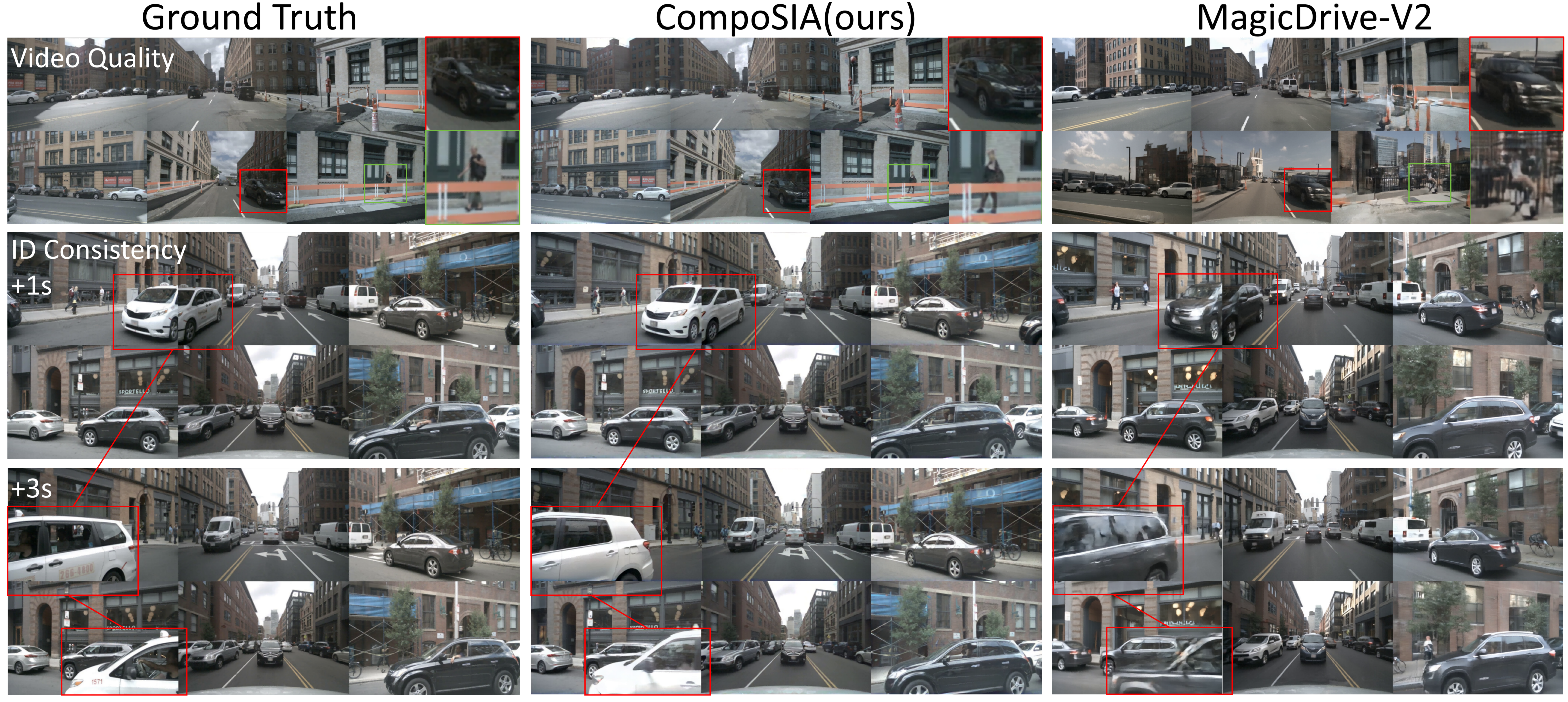}
\captionof{figure}{\textbf{Comparison of generation quality.} \modelname renders vehicles and pedestrians with more structurally coherent geometry, while maintaining stronger cross-view and temporal consistency.}
\label{fig:exp_0}
\vspace{-6ex}
\end{table}

\subsection{Experimental Setups}

\noindent\textbf{Implementation Details.}
Our \modelname is initialized from Wan2.1-T2V-1.3B~\cite{wan2025wan}.
Considering the fast temporal dynamics in driving scenarios, we fine-tune a version of the Wan VAE without temporal downsampling, setting the temporal stride to 1 instead of the original $4\times$ compression. 
For the DiT module, we additionally introduce a small projector to inject structure condition. 
Similar projectors are also added for achieving the low-dimensional PRoPE attention and the residual modulation for ego-action conditioning.
We train \modelname on 16 NVIDIA 80GB A100 GPUs, with learning rate $2\times10^{-4}$ for the action projector
and $1\times10^{-5}$ for all other components,
The weight decay of $5\times10^{-2}$ is applied to all parameters.
For training of identity injection, we replace latents with identity image only at high noise levels ($t > 0.2$ in~\cref{equ:flow_matching}),
to avoid disrupting the denoising path at low noise levels for high-quality video generation.

\noindent\textbf{Dataset.}
We train \modelname on nuScenes~\cite{caesar2020nuscenes} and our self-collected driving data.
For nuScenes, we use 700 multi-view videos of 20 seconds for training and 150 for validation.
To cover a wider range of element types, more diverse scenes, and more complex high-speed driving cases,
we further incorporate 100 hours of internal, multi-view autonomous driving data for training.
A mixed-resolution training strategy is applied using both $33 \times 256 \times 512$ and $33 \times 480 \times 960$ clips 
to facilitate faster convergence of the condition modeling.
All data are trained at 10 Hz.

\subsection{Comparisons of Generation Quality}
For video generation, we follow benchmarks from~\cite{chen2025eccv}, utilizing FVD for evaluating of video quality.
We further report VBench Score~\cite{huang2024vbench} (Quality Score only) for a more comprehensive and multi-dimensional evaluation.
The results are presented in~\cref{tab:FVD} and~\cref{fig:exp_0}.
\modelname renders vehicles and pedestrians with higher fidelity, 
while maintaining cross-view and temporal consistency.

\subsection{Structure and Identity Control}
\begin{figure}[t]
\centering
\includegraphics[width=1\linewidth]{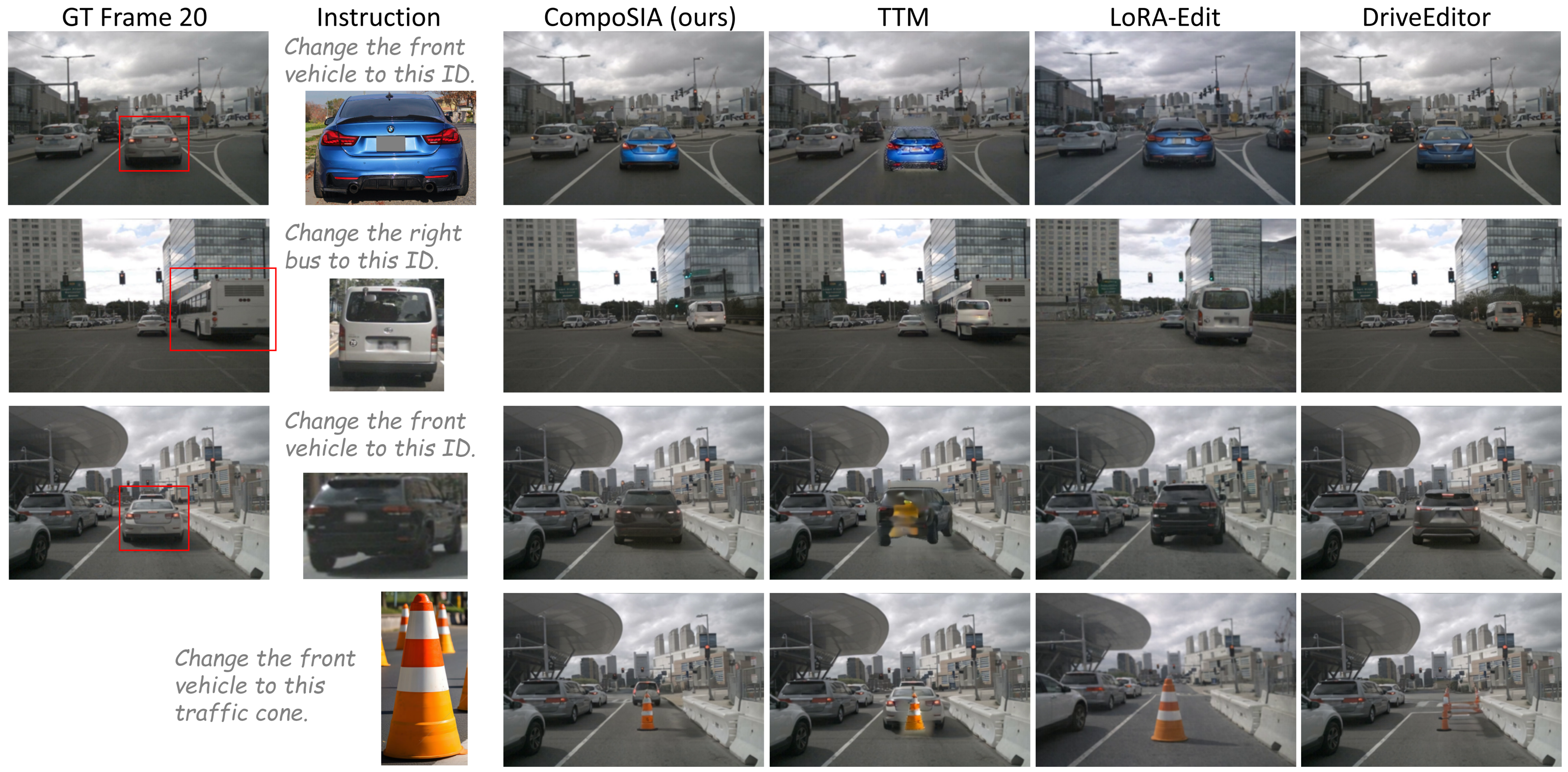}
\vspace{-2ex}
\caption{\textbf{Comparison on identity control}. \modelname enables precise identity injection across diverse element poses, categories, and in-the-wild reference images.}
\label{fig:exp_2}
\vspace{-3ex}
\end{figure}

For identity control shown in~\cref{fig:exp_2}, the measured video quality is reported in~\cref{tab:FVD}.
We apply TTM~\cite{singer2025time} as a training-free sampling strategy on a model trained on our driving data, which fails to generalize to reference images captured from novel element poses.
LoRA-Edit~\cite{gao2025lora} achieves high fidelity via first-frame editing but cannot precisely control the insertion location of the generated element.
DriveEditor~\cite{liang2025driveeditor} injects reference images through a conditional branch, yet the transferred identity deviates noticeably from the original.
In contrast, \modelname supports editing diverse elements with controllable positions and orientations, across heterogeneous categories, while maintaining strong identity fidelity to reference images.

\begin{wraptable}{r}{0.6\textwidth}
\vspace{-3ex}
\caption{\textbf{Evaluation of action control.} $\mathtt{TransErr}$ is $1000\times$.}
\label{tab:traj}
\centering
\begin{adjustbox}{width=0.6\textwidth}
\begin{tabular}{
>{\arraybackslash}p{0.25\textwidth}
>{\centering\arraybackslash}p{0.15\textwidth} 
>{\centering\arraybackslash}p{0.15\textwidth}
>{\centering\arraybackslash}p{0.15\textwidth}
>{\centering\arraybackslash}p{0.15\textwidth}
}
\toprule
\multirow{2}{*}{\textbf{Method}} 
& \multicolumn{2}{c}{\textbf{Action Following}} 
& \multicolumn{2}{c}{\textbf{Action Editing}} \\
& $\mathtt{RotErr}$ $\downarrow$ 
& $\mathtt{TransErr}$ $\downarrow$
& $\mathtt{RotErr}$ $\downarrow$ 
& $\mathtt{TransErr}$ $\downarrow$
\\
\midrule
ReCamMaster~\cite{bai2025recammaster} & 1.12 & 20.35 & 2.17 & 25.45\\
Vista~\cite{gao2024vista} & 0.81 & 14. 25 & 2.33 & 28.12\\
MagicDrive-V2~\cite{gao2025magicdrive} & 0.76 & 13.66 & 2.21 & 22.86\\
\rowcolor{gray!15}
\modelname~(Ours) & \textbf{0.55} & \textbf{7.37} & \textbf{1.54} & \textbf{12.15}\\
\bottomrule
\end{tabular}
\end{adjustbox}
\vspace{-6ex}
\end{wraptable}

\subsection{Action Control}

\begin{figure}[t]
\centering
\includegraphics[width=1\linewidth]{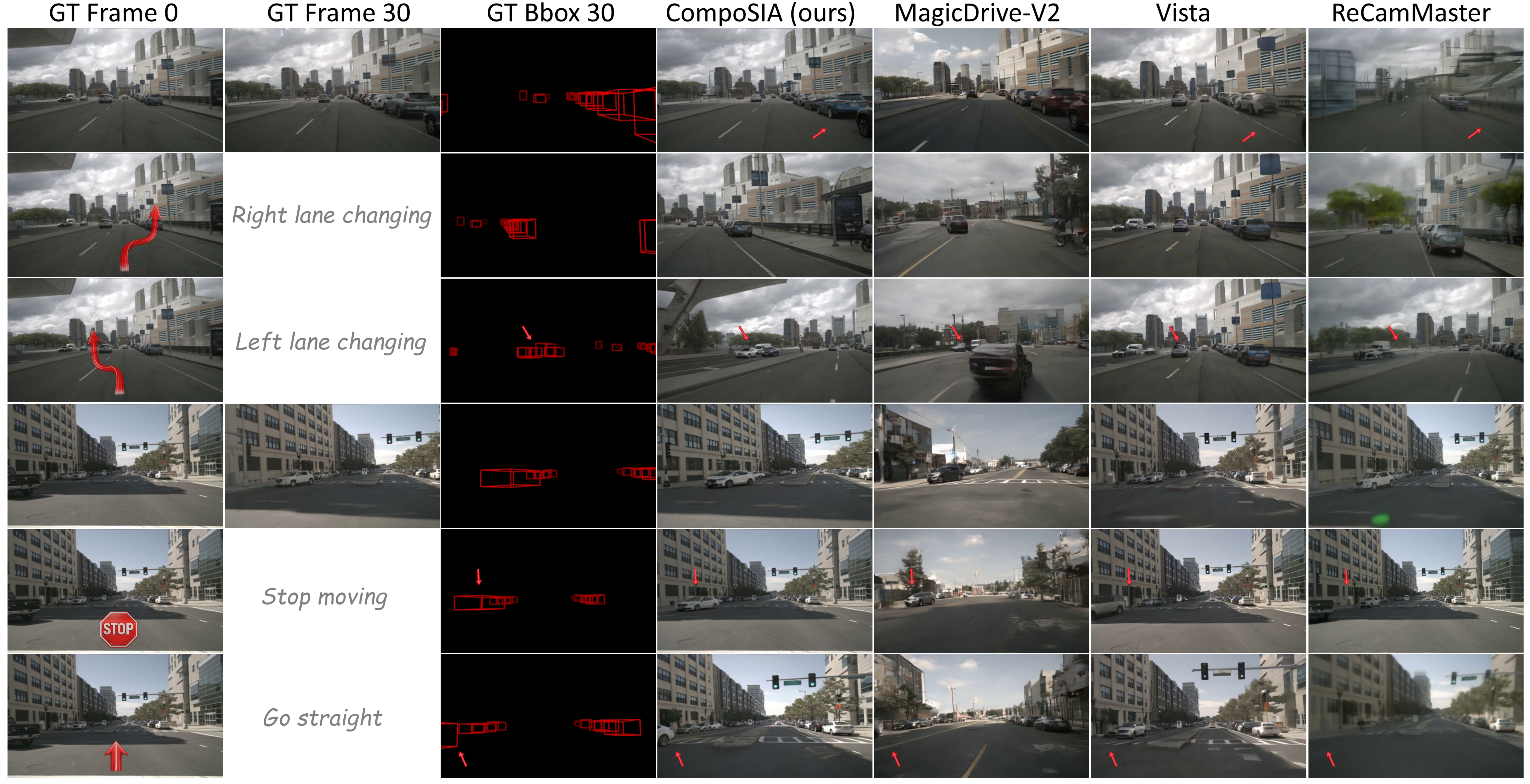}
\caption{\textbf{Comparison on action control}. \modelname faithfully executes diverse action commands with precise frame-wise controllability.}
\label{fig:exp_1}
\vspace{-4ex}
\end{figure}

In~\cref{tab:traj}, we evaluate the action control by estimating trajectories
from the generated videos using VGGT~\cite{wang2025vggt},
aligning the predicted trajectories with the ground-truth trajectories.
We report
\begin{equation}
\begin{aligned}
\mathtt{RotErr} &= \sum_{i=1}^{n} 
\arccos \frac{\operatorname{tr}(\mathbf{R}_{gen}^{i}\mathbf{R}_{gt}^{i\top}) - 1}{2}, \\
\mathtt{TransErr} &= \sum_{i=1}^{n} 
\|\mathbf{T}_{gt}^{i} - \mathbf{T}_{gen}^{i}\|_2^2 .
\end{aligned}
\end{equation}
where $\mathbf{R}_{gen}^{i}$ and $\mathbf{R}_{gt}^{i}$ denote the estimated and ground-truth rotation matrices, and $\mathbf{T}_{gen}^{i}$ and $\mathbf{T}_{gt}^{i}$ denote the corresponding translation vectors, all at the $i$-th frame.
We compare MagicDrive-V2~\cite{gao2025magicdrive} (trajectory embedding), Vista~\cite{gao2024vista} (trajectory cross-attention), and ReCamMaster~\cite{bai2025recammaster} (3D attention) under the same first-frame image condition.
For each method, we modify the action to perform different maneuvers, including left lane change, right lane change, stop, and accelerating straight.
As shown in~\cref{fig:exp_1}, only our approach accurately follows the frame-wise action sequence.

\subsection{Ablation Study}

\begin{figure}[t]
\centering
\includegraphics[width=1\linewidth]{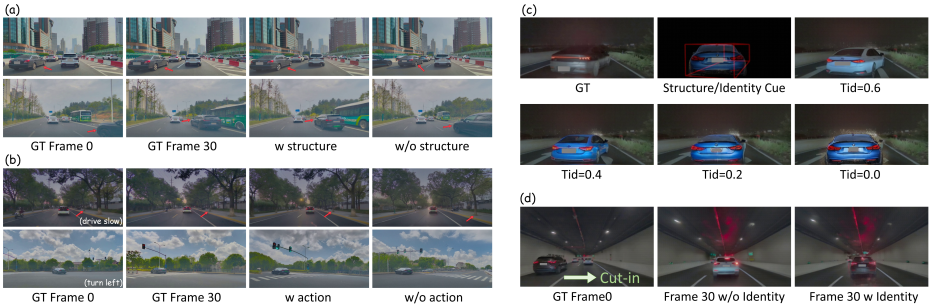}
\caption{\textbf{Ablation studies on structure, action, and identity injection.}(a) Structure ablation.(b) Action ablation.(c) Effect of $T_{\text{id}}$.(d) Identity injection improves cross-frame consistency under complex illumination.}
\label{fig:exp_5}
\vspace{0ex}
\end{figure}

\noindent\textbf{Ablating Structure Control.}
To demonstrate the role of structure control, we disable structure conditioning while retaining action guidance during sampling in~\cref{fig:exp_5} (a).
As a result, surrounding vehicles fail to maintain consistent motion and spatial alignment.

\noindent\textbf{Ablating Action Control.}
To verify that action control is effective rather than ego motion being leaked from structure cues, we remove the action branch during sampling.
In~\cref{fig:exp_5} (b), the structure conditioning are kept identical to the ground truth while no action control is provided, 
Without which the ego motion becomes unstable.
We further ablate different action injection strategies (r.m. for residual modulation and p.a. for PRoPE attention) in~\cref{tab:ablate_action} to study their impact on controllability, showing that our hierarchical dual-branch design achieves the best performance.

\begin{wraptable}{r}{0.4\textwidth}
\vspace{-8ex}
\caption{\textbf{Ablation of action conditioning.} $\mathtt{TransErr}$ is $1000\times$.}
\label{tab:ablate_action}
\centering
\begin{adjustbox}{width=0.4\textwidth}
\begin{tabular}{
>{\arraybackslash}p{0.15\textwidth}
>{\centering\arraybackslash}p{0.15\textwidth}
>{\centering\arraybackslash}p{0.15\textwidth}
}
\toprule
\multirow{2}{*}{\textbf{Condition}} 
& \multicolumn{2}{c}{\textbf{Action Following}} \\
& $\mathtt{RotErr}$ $\downarrow$ 
& $\mathtt{TransErr}$ $\downarrow$
\\
\midrule
w/o r.m. & 2.84 & 15.80\\
w/o p.a. & 0.62 & 11.24\\
\rowcolor{gray!15}
Full & \textbf{0.55} & \textbf{7.37}\\
\bottomrule
\end{tabular}
\end{adjustbox}
\vspace{-3ex}
\end{wraptable}

\noindent\textbf{Design of $T_{\text{id}}$ for Identity Control.}
We further ablate the effect of different $T_{\text{id}}$, the stopping step of identity injection, which directly influences the balance between editing freedom and identity preservation.
As can be found in~\cref{fig:exp_5} (c), earlier stopping ($T_{\text{id}}=0.6$) increases generation freedom but reduces similarity to the reference image, whereas later stopping ($T_{\text{id}}=0.2$) enforces stronger identity preservation but increasingly anchors the generation to the reference image.
Nevertheless, since identity cues are injected during training across a wide range of noise levels ($t>0.2$), the stopping step remains robust for different elements.
Therefore, no per-case tuning is required, and we adopt $T_{\text{id}} = 0.4$ as a practical trade-off.

\noindent\textbf{Effect of Identity Cue on ID Preservation.}
Beyond its role in controllable editing, identity cue injection also improves ID stability during generation.
As shown in~\cref{fig:exp_5} (d), in scenarios with complex illumination transitions (\eg, driving through tunnels), relying solely on the first-frame reference leads to inconsistent scene element appearances across frames.
In contrast, injecting identity cues throughout the denoising process substantially reduces identity drift.
This noise-level identity conditioning offers a principled and temporally coherent mechanism for stabilizing identity without sacrificing generative flexibility.

\subsection{Evaluation of Planning Robustness}
We conduct planning robustness evaluation on nuScenes~\cite{caesar2020nuscenes}.
Specifically, we synthesize multiple groups of data.
The first group strictly follows the original structure, identity and action, serving to isolate the impact of potential video quality degradation on planner decisions.
Other groups edit both surrounding agents and ego actions,
together with identity replacement, to construct more challenging scenarios.
These are designed to demonstrate that our synthesized data represent long-tail and adversarial cases for planning.
We adopt Epona~\cite{zhang2025epona} as the end-to-end planner and evaluate all groups in an open-loop setting, reporting L2 distance and collision rate based on the predicted trajectories.
\cref{tab:epona} shows noticeable metric degradation relative to the generated baseline, indicating that the introduced edits increase task complexity and challenge the planner's generalization capability.
Detailed editing modalities can be found in the supplementary materials.

\begin{table}[t]
\caption{\textbf{End-to-end motion planning performance on nuScenes~\cite{caesar2020nuscenes}.}}
\label{tab:epona}
\vspace{-2mm}
\centering
\scalebox{0.8}{
\begin{tabular}{
>{\arraybackslash}p{0.25\textwidth}
>{\centering\arraybackslash}p{0.07\textwidth} 
>{\centering\arraybackslash}p{0.07\textwidth}
>{\centering\arraybackslash}p{0.07\textwidth}
>{\arraybackslash\columncolor{gray!15}}p{0.15\textwidth}
>{\centering\arraybackslash}p{0.07\textwidth} 
>{\centering\arraybackslash}p{0.07\textwidth}
>{\centering\arraybackslash}p{0.07\textwidth}
>{\arraybackslash\columncolor{gray!15}}p{0.15\textwidth}
}
\toprule
\multirow{2}{*}{\textbf{Type}} 
& \multicolumn{4}{c}{\textbf{L2(m) $\downarrow$}} 
& \multicolumn{4}{c}{\textbf{Collision Rate(\%) $\downarrow$}} \\

& $\mathtt{1s}$ 
& $\mathtt{2s}$
& $\mathtt{3s}$
& $\mathtt{Avg.}$
& $\mathtt{1s}$ 
& $\mathtt{2s}$
& $\mathtt{3s}$
& $\mathtt{Avg.}$
\\
\midrule
Following GT
& 0.68 & 1.33 & 2.24 & 1.42 & 0.04 & 0.24 & 0.76 & 0.35\\
Following Generation
& 0.72 & 1.41 & 2.83 & \underline{1.65} & 0.08 & 0.36 & 1.32 & \underline{0.59}\\
Editing Structure
& -- & -- & -- & -- & 0.72 & 2.68 & 5.28 & 2.89 {\scriptsize\color{red}$\uparrow$390\%}\\
Editing Identity
& 0.89 & 1.77 & 3.91 & 2.19 {\scriptsize\color{red}$\uparrow$33\%} & 0.12 & 0.48 & 1.64 & 0.75 {\scriptsize\color{red}$\uparrow$27\%}\\
Editing Action
& 0.91 & 1.83 & 4.22 & 2.32 {\scriptsize\color{red}$\uparrow$41\%} & 0.16 & 0.76 & 2.64 &1.19 {\scriptsize\color{red}$\uparrow$102\%}\\

\bottomrule
\end{tabular}
}
\vspace{-6mm}
\end{table}

\section{Limitations}
\label{sec:Limitations}

Despite its effectiveness, \modelname has some limitations.
First, our training data primarily consists of driving scenes, which restricts identity editing generalization to total out-of-distribution element categories (\eg, animals).
Although our identity formulation does not explicitly constrain category type, broader generalization would benefit from scaling training data to include more diverse video sources.
Second, our current identity editing pipeline requires specifying an approximate 3D bounding box size (\ie, height, width, length) for the reference image during replacement or insertion.
At present, this estimation is facilitated by a large language model (\eg, Gemini~\cite{gemini2024}), but the process remains semi-automatic.
Future work may explore more user-friendly editing interfaces with automatic spatial grounding and interactive refinement.

\section{Conclusion}
\label{sec:Conclusion}

In this paper, we present \modelname, a compositional driving world simulator to generate adversarial Scenario through unified and disentangled control.
By introducing factor-specific conditioning branches, our framework enables both independent and joint manipulation of scene layout, object identity, and ego actions.
Extensive experiments demonstrate that \modelname consistently surpasses prior driving world models, driving-specific editing approaches, and general controllable generation frameworks in both generation fidelity and condition alignment.
Building upon these capabilities, we construct controllable adversarial scenarios for systematic planner stress-testing, uncovering robustness limitations that remain hidden under conventional benchmarks.
We believe \modelname advances controllable driving world modeling and provides a principled foundation for reliable scenario synthesis and robustness auditing in autonomous driving systems.

%
%
\bibliographystyle{splncs04}
\bibliography{main}

\clearpage
\appendix

\begin{center}
\begin{minipage}{1.0\linewidth}
\includegraphics[width=1.0\linewidth]{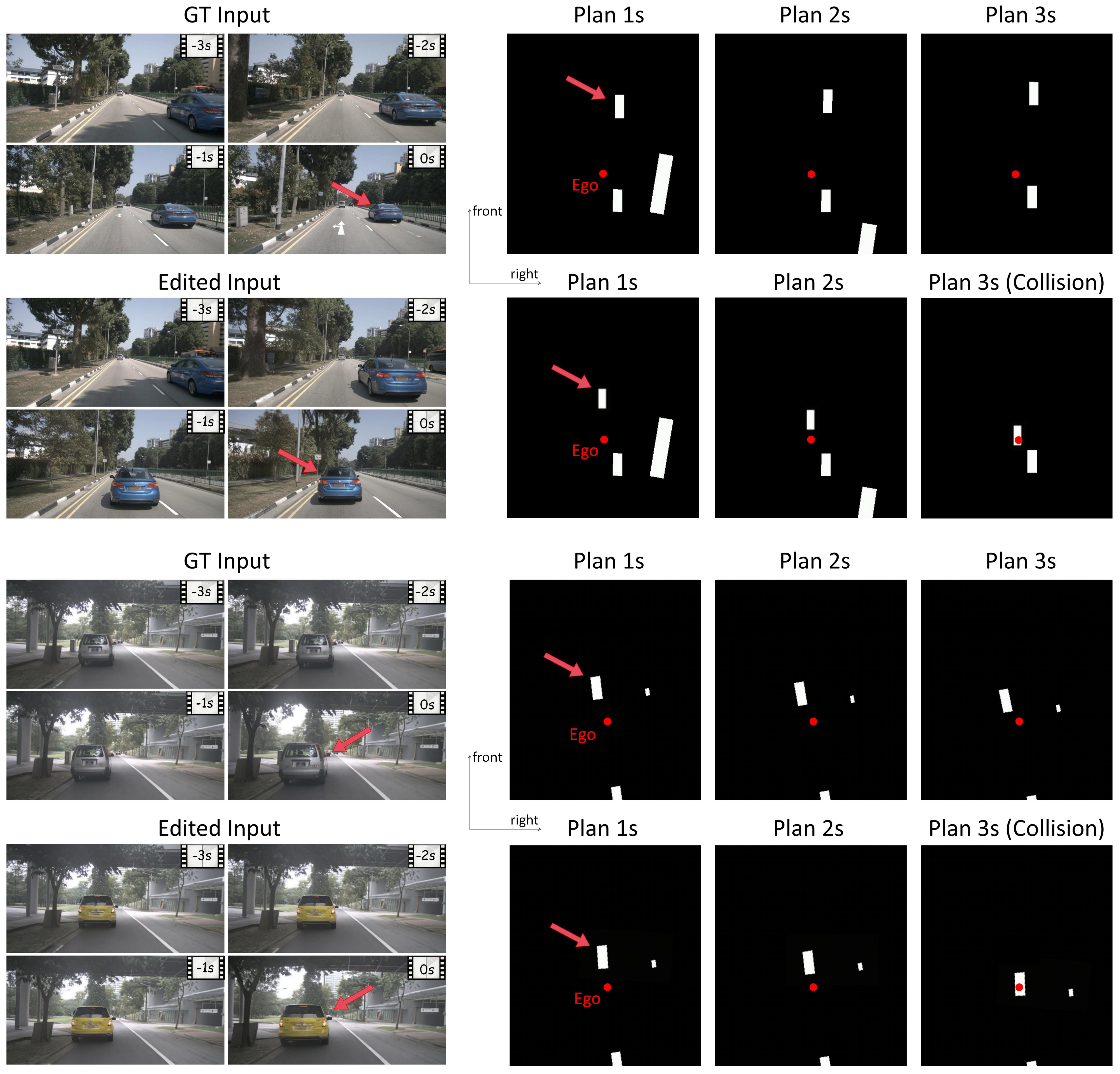}
\vspace{-4ex}
\captionof{figure}{\textbf{Edited historical observations induce planner failure.} \textbf{Left:} Historical frames provided to Epona~\cite{zhang2025epona}. Rows 1 and 3 show the original ground-truth history, whereas rows 2 and 4 present the edited counterparts. We illustrate two representative cases: vehicle cut-in and identity replacement. \textbf{Right:} Predicted BEV planning maps at $\mathtt{1s}$, $\mathtt{2s}$, and $\mathtt{3s}$ into the future. Given the original history (rows 1 and 3), the planner generates safe future trajectories. In contrast, when conditioned on the edited history (rows 2 and 4), a collision occurs at 3 s, showing that the edited scenarios can effectively expose planner vulnerabilities.}
\label{fig: supp_plan}
\end{minipage}
\end{center}
\setlength{\parindent}{2em}

\section{Detailed Editing Modalities for Planning Evaluation}
\label{sec:Detailed Editing Modalities for Planning Evaluation}

\noindent\textbf{Structure Editing.}
For structure-level intervention, we construct two representative traffic editing modes through automatic scenario filtering.
The first mode targets cut-in events.
We automatically identify scenes where the ego vehicle is moving forward, no leading vehicle exists in the current lane, and at least one vehicle is present in the adjacent lane.
In these scenes, the adjacent vehicle is edited to cut into the ego lane, forming a lateral intrusion scenario.
This category accounts for 21.6\% of all candidate clips.
The second mode targets front-vehicle insertion.
We automatically identify scenes where the ego vehicle is moving forward with no leading vehicle ahead, and insert a new vehicle at a distant frontal position along the ego lane.
This category accounts for 43.9\% of all candidate clips.
For both modes, the edited segments replace the corresponding portions in the full sequences of following generation.
Since no ground-truth future trajectory exists for edited scenarios, we evaluate planning robustness using average collision rate under planner rollouts.

\begin{figure}[t]
\centering
\includegraphics[width=1\linewidth]{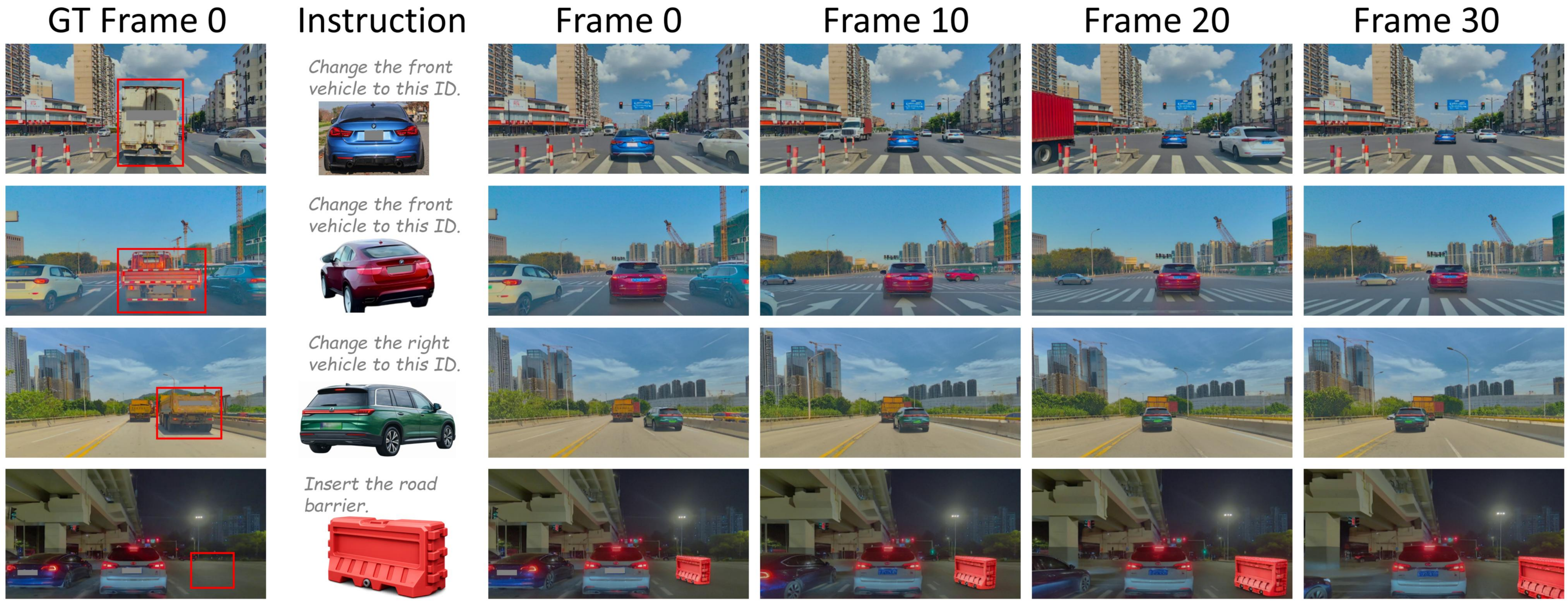}
\caption{\textbf{More examples of identity editing.} Given a single in-the-wild reference image of the target object, our method replaces the identity of a scene element while preserving the original scene structure and motion. Despite large variations in element pose, categories, and appearance between the reference image and the generated sequence, the edited object is rendered with temporally consistent and realistic identity.}
\label{fig:supp_1}
\end{figure}

\noindent\textbf{Identity Editing.}
For identity-level intervention, we select scenes where a leading vehicle is present in front of the ego vehicle and replace its visual identity using a single in-the-wild reference image, while keeping the original structure condition unchanged.
This category accounts for 36.3\% of all candidate clips.
Because the underlying structure remains unchanged, the original ground-truth future trajectory is still valid for evaluation.
We therefore report both trajectory prediction error with respect to ground truth and future collision rate under planner rollouts.

\noindent\textbf{Action Editing.}
For action-level intervention, we apply constrained perturbations to the ego action sequence while keeping structure and identity conditions unchanged.
Specifically, we fix the start and end positions and impose boundary constraints on velocity and acceleration.
A smooth bump-shaped curve is then introduced in the middle portion of the action sequence to perturb ego motion while preserving temporal continuity.
This design produces physically plausible action variations and allows reuse of the original future trajectory for evaluation.
We apply action editing to all clips involving ego motion, accounting for 74.2\% of all candidate clips.
Because surrounding scene conditions remain unchanged, both trajectory prediction error and future collision rate are reported.

The reported proportions are computed independently for each editing modality, based on separately selected candidate scenes, and therefore are not mutually exclusive.
\cref{fig: supp_plan} presents a representative example in which a structure-level edit leads to planner failure.
More broadly, Tab. \textcolor{red}{5} in the main paper shows that all editing modalities cause substantial degradation in planning performance compared with the generated baseline, further highlighting the importance of corner-case generation for stress-testing autonomous driving planners.

\begin{figure}[t]
\centering
\includegraphics[width=1\linewidth]{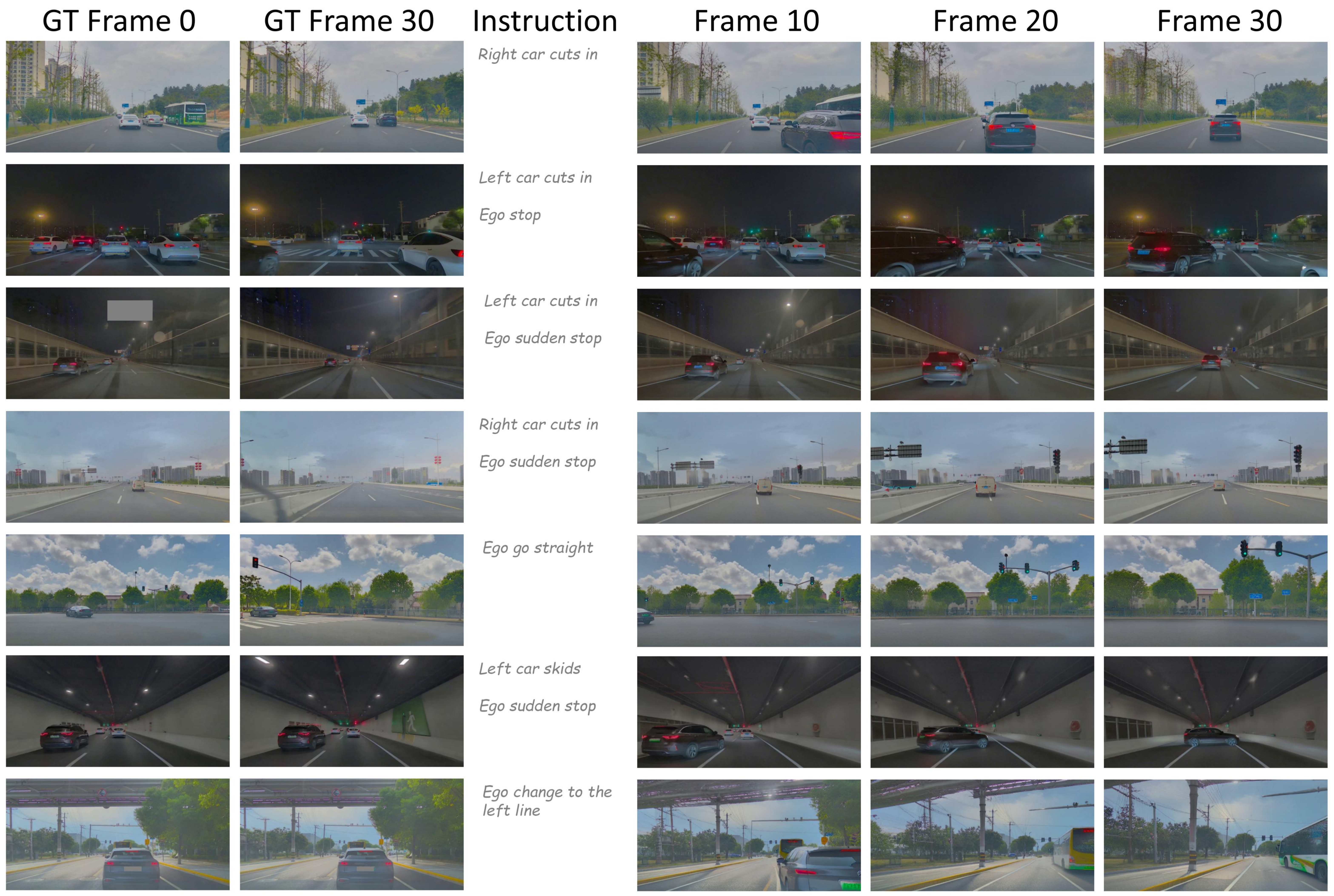}
\caption{\textbf{Additional results on compositional structure-and-action control.} We present more examples that jointly modify scene structure and ego action, including cut-in vehicles, ego stopping, sudden braking, lane changes, and skidding events. The generated videos remain temporally coherent while following the intended compositional interventions across diverse traffic scenes. The instruction column is included only for human-readable description of the applied edits; no real text prompt is used during the editing process.}
\label{fig:supp_2}
\end{figure}

\section{Modified VAE for Driving Video Modeling}
\label{sec:Additional Implementation Details}

Compared with general video domains, driving scenes exhibit substantially larger motion displacement and appearance variation between adjacent frames under the same frame rate, making aggressive temporal compression less suitable for preserving controllable dynamics.

We adopt the pretrained Wan-VAE from Wan2.1-T2V-1.3B~\cite{wan2025wan} as the initialization of our latent video tokenizer.
The original Wan-VAE applies a spatio-temporal compression ratio of $4\times8\times8$, where temporal downsampling is performed through an intermediate temporal compression stage.
For driving video generation, we remove the temporal compression layer and set the temporal stride to 1, while keeping the spatial compression unchanged.
This modification preserves frame-level temporal resolution in latent space, which is critical for driving scenarios containing fast motion changes such as steering, braking, and nearby object interactions.

Starting from the pretrained Wan-VAE weights, we fine-tune the modified VAE on our 100h self-collected driving dataset for 7 days using 8 NVIDIA 80GB A100 GPUs.
Since removing the temporal compression stage substantially changes the internal feature distribution across the encoder-decoder pipeline, we fine-tune all VAE modules.
We adopt the same training losses, learning rate, and optimization schedule as the original Wan-VAE training protocol; detailed settings can be found in~\cite{wan2025wan}.

\section{Optimization Details}
\label{sec:Optimization Details}

We perform full supervised fine-tuning of the diffusion DiT backbone, starting from the pretrained Wan2.1-T2V-1.3B initialization.
The backbone is optimized with a learning rate of $1\times10^{-5}$ and a linear warm-up of 500 steps.
A weight decay of $5\times10^{-2}$ is applied to all parameters.
For the action modulation branch, we adopt a larger learning rate of $2\times10^{-4}$ to facilitate stable adaptation of the newly introduced control layers.
Optimization is performed using AdamW~\cite{loshchilov2017fixing} with $\epsilon=10^{-10}$ and gradient clipping with maximum norm 0.05.
Training follows the standard flow-matching formulation with 1000 discrete noise levels using the FlowMatchEulerDiscreteScheduler~\cite{esser2024scaling}.
The entire training takes around 4 days for 20,000 steps on 16 NVIDIA 80GB A100 GPUs, with per-GPU batch size 4.

\section{Inference Details}
\label{sec:Inference Details}

Unless otherwise specified, all videos are generated using 50 denoising steps under the FlowMatchEulerDiscreteScheduler~\cite{esser2024scaling}.
During inference, identity injection replaces target-region latents at each denoising step while $t > T_{\text{id}} = 0.4$.

\section{More Qualitative Results}
\label{sec:More Qualitative Results}

In~\cref{fig:supp_1} and~\cref{fig:supp_2} We provide additional qualitative results to further illustrate the controllability and temporal coherence of our method under diverse editing settings, including structure, identity, and action interventions, as well as their compositions in challenging driving scenarios.

\end{document}